\documentclass[journal,twoside,web]{ieeecolor}
\usepackage{tmi}
\usepackage{cite}
\usepackage{amsmath,amssymb,amsfonts}
\usepackage{algorithmic}
\usepackage{graphicx}
\usepackage{textcomp}
\usepackage{multirow}
\usepackage{pifont}
\usepackage{url}
\newcommand{\xmark}{\ding{55}}
\def\BibTeX{{\rm B\kern-.05em{\sc i\kern-.025em b}\kern-.08em
    T\kern-.1667em\lower.7ex\hbox{E}\kern-.125emX}}
\begin{document}
\title{An Anisotropic Cross-View Texture Transfer with Multi-Reference Non-Local Attention for CT Slice Interpolation}
\author{Kwang-Hyun Uhm, Hyunjun Cho, Sung-Hoo Hong, and Seung-Won Jung, \IEEEmembership{Senior Member, IEEE} 
\thanks{This work was supported by Artificial intelligence industrial convergence cluster development project funded by the Ministry of Science and ICT(MSIT, Korea)\&Gwangju Metropolitan City and was supported by the Technology development Program (RS-2024-00460263) funded by the Ministry of SMEs and Startups (MSS, Korea). (\textit{Corresponding author: Seung-Won Jung.)}}
\thanks{Kwang-Hyun Uhm is with the Department of Artificial Intelligence, Gachon University, Gyeonggi-do 13120, Republic of Korea (e-mail: khuhm@gachon.ac.kr).}
\thanks{Hyunjun Cho is with the Department of Electrical Engineering, Korea University, Seoul 02841, South Korea, and with MedAI, Seoul 02841, Republic of Korea (e-mail: chohj1111@korea.ac.kr).}
\thanks{Seung-Won Jung is with the Department of Electrical Engineering, Korea University, Seoul 02841, Republic of Korea (e-mail: swjung83@korea.ac.kr).}
\thanks{Sung-Hoo Hong is with the Department of Urology, The Catholic University of Korea, Seoul 07442, Republic of Korea. (e-mail: toomey@catholic.ac.kr )}}

\maketitle

\begin{abstract}
Computed tomography (CT) is one of the most widely used non-invasive imaging modalities for medical diagnosis.
In clinical practice, CT images are usually acquired with large slice thicknesses due to the high cost of memory storage and operation time, resulting in an anisotropic CT volume with much lower inter-slice resolution than in-plane resolution.
Since such inconsistent resolution may lead to difficulties in disease diagnosis, deep learning-based volumetric super-resolution methods have been developed to improve inter-slice resolution. Most existing methods conduct single-image super-resolution on the through-plane or synthesize intermediate slices from adjacent slices; however, the anisotropic characteristic of 3D CT volume has not been well explored.
In this paper, we propose a novel cross-view texture transfer approach for CT slice interpolation by fully utilizing the anisotropic nature of 3D CT volume.
Specifically, we design a unique framework that takes high-resolution in-plane texture details as a reference and transfers them to low-resolution through-plane images. To this end, we introduce a multi-reference non-local attention module that extracts meaningful features for reconstructing through-plane high-frequency details from multiple in-plane images.
Through extensive experiments, we demonstrate that our method performs significantly better in CT slice interpolation than existing competing methods on public CT datasets including a real-paired benchmark, verifying the effectiveness of the proposed framework. The source code of this
work is available at https://github.com/khuhm/ACVTT.
\end{abstract}


\begin{figure*}[t]
\centering
\includegraphics[width=0.9\textwidth]{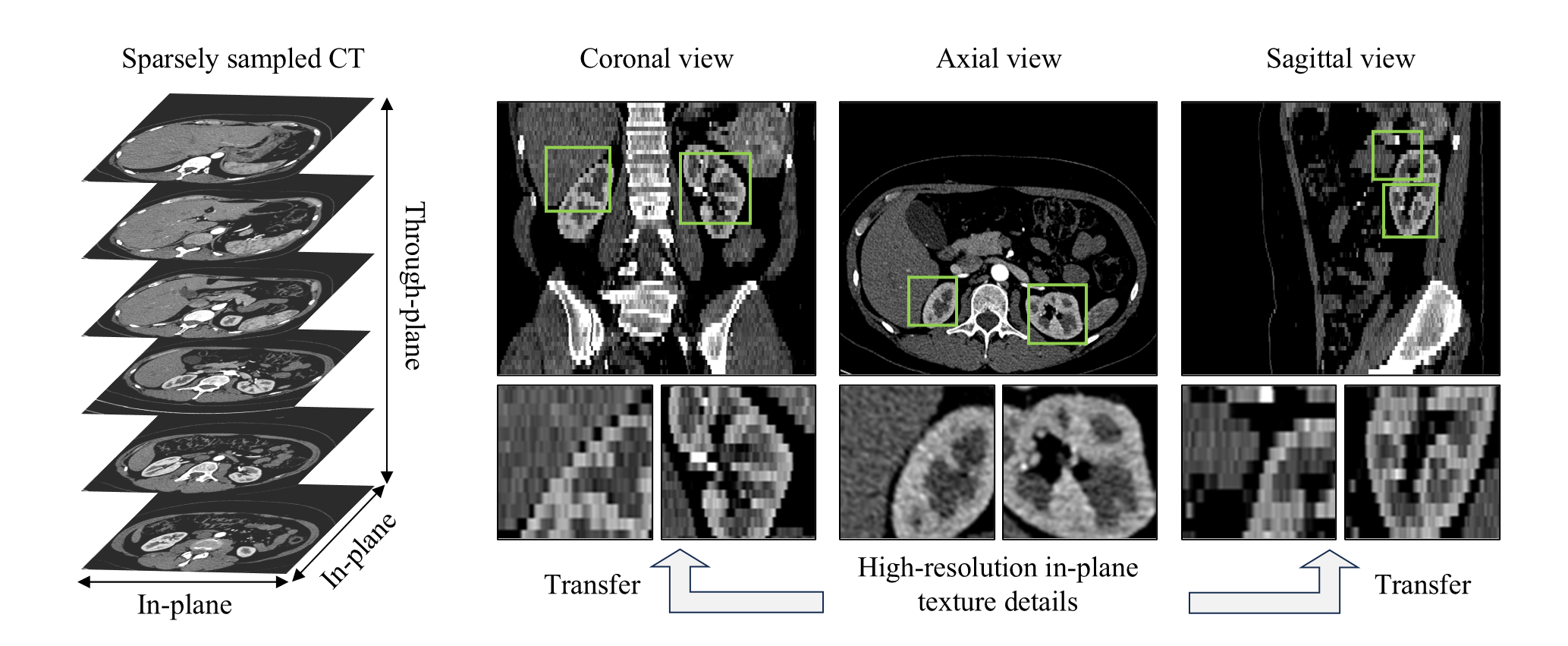}
\caption{Our cross-view texture transfer for slice interpolation of anisotropic CT. In sparsely sampled CT, the resolution of through-plane (coronal and sagittal views) images is much lower than that of in-plane (axial view) images. We propose to transfer the high-resolution in-plane texture details to the low-resolution through-plane images to restore the high-frequency details of through-plane images.
} \label{fig1}
\end{figure*}

\section{Introduction}
\label{sec:introduction}
\IEEEPARstart{R}{ecently}, medical imaging techniques such as computed tomography (CT) and magnetic resonance imaging (MRI) have played a crucial role in providing detailed visual information about the internal structures of the human body~\cite{haaga2016computed, imaging_rcc}.
In particular, CT is widely used for diagnosis due to its lower cost and shorter imaging time compared to MRI~\cite{pic_review, black2014cost}.
Despite its advantages, CT imaging is often compromised by the trade-off between image quality and practical constraints such as radiation dose and data storage. 
Large slice thicknesses, often used to reduce scan time and radiation exposure, result in anisotropic 3D volumes, where the through-plane (coronal and sagittal) resolution is substantially lower than the in-plane (axial) resolution, as illustrated in Fig. 1. 
The anisotropic resolution of CT images presents several challenges in clinical practice and medical research~\cite{TVSRN, I3Net}. For instance, the lower through-plane resolution can obscure subtle lesions or anatomical features, complicating the diagnostic process.
Additionally, this inconsistency poses significant hurdles for advanced image processing tasks, such as 3D reconstruction, segmentation, and analysis, particularly affecting the accuracy in identifying structures during segmentation and in reconstructing 3D models from 2D slices.
As a result, CT slice interpolation has emerged as a practical solution.

In response to these challenges, researchers have increasingly turned to advanced image enhancement techniques, particularly those based on deep learning, to improve the quality and diagnostic utility of CT images~\cite{SAINT, DA-VSR, Incremental, TVSRN}. Volumetric super-resolution methods aim to address the resolution gap by synthesizing high-resolution 3D images from lower-resolution scans. These techniques leverage the capabilities of deep neural networks to learn complex mappings between low-resolution and high-resolution images, effectively reconstructing missing details and improving image clarity.
Current methods for CT slice interpolation typically focus on single-image super-resolution~\cite{MDSR, Meta-SR, RDN}, enhancing the resolution of individual slices in the through-plane direction~\cite{SAINT, DA-VSR}, or generating intermediate slices by synthesizing information from adjacent slices~\cite{Incremental, TVSRN}. While these approaches have shown promise, they often fall short of leveraging the full potential of the anisotropic information embedded within the 3D volume. By primarily focusing on through-plane slices, these methods may neglect the rich texture details present in high-resolution in-plane slices, which could be crucial for achieving superior interpolation results.

On the other hand, another line of research, reference-based super-resolution (RefSR) techniques have emerged as a powerful approach to enhance the resolution of images by utilizing additional reference images that provide high-frequency details~\cite{CrossNet,SRNTT,yan2020towards,C2Matching,RZSR,LMR}. These methods involve matching features between low-resolution target images and high-resolution reference images, followed by the transfer of fine textures based on the matched correspondences.
This has motivated us to leverage the high-resolution in-plane details as a reference to more effectively enhance the low-resolution through-plane images. This requires innovative models that can integrate information from multiple views and capture the complex relationships between different planes within a volume.

In this paper, we propose the Anisotropic Cross-View Texture Transfer Network (ACVTT) for CT slice interpolation by fully exploiting the anisotropic nature of 3D CT volumes. Our framework creatively leverages multiple high-resolution in-plane slices from the same CT scan as intra-volume references to enhance low-resolution through-plane images, eliminating the need for external datasets and ensuring high contextual relevance.
Motivated by the observation that local anatomical structures often exhibit consistent patterns across different views (as illustrated in Fig.~\ref{fig1}), we design a novel cross-view texture transfer strategy that extracts fine in-plane textures and transfers them to through-plane slices.
To effectively utilize multiple references, we introduce the Multi-Reference Non-Local Attention (MRNLA) module, which selectively aggregates high-frequency features across multiple in-plane slices for accurate through-plane reconstruction.
During training, our model is supervised with paired high-resolution and low-resolution CT volumes. However, during inference, it exclusively utilizes intra-volume axial slices as references without requiring any external datasets, ensuring high contextual relevance to the low-resolution through-plane images.
Extensive experiments on three public CT datasets, including a real thick-thin paired CT benchmark, demonstrate that ACVTT consistently outperforms existing methods in slice interpolation tasks, verifying the effectiveness and robustness of our approach.

\begin{figure*}[t]
\centering
\includegraphics[width=0.8\textwidth]{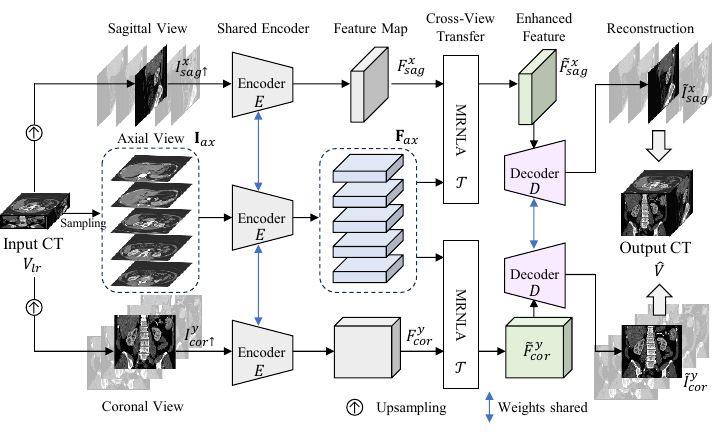}
\caption{Overall framework of our proposed ACVTT. It first takes a sparsely sampled input CT volume and extracts features from multiple planes (axial, coronal, and sagittal views) through a shared encoder. Then, axial slice features are utilized as references to enhance through-plane images via our cross-view texture transfer mechanism. To effectively transfer multiple high-resolution in-plane features, the MRNLA is newly introduced. The reconstructed volumes from coronal and sagittal views are combined to produce a final densely sampled 3D CT volume.
} \label{fig2}
\end{figure*}

\section{Related Work}
\subsection{CT slice interpolation}
Various deep learning-based methods have been proposed for CT slice interpolation and through-plane super-resolution.
Early approaches, such as SAINT~\cite{SAINT} and DA-VSR~\cite{DA-VSR}, employed CNN-based architectures to enhance through-plane views, using supervised or self-supervised signals based on resolution differences across planes.
Self-supervised frameworks have further evolved, with methods like Incremental Distillation~\cite{Incremental} and SR4ZCT~\cite{sr4zct} leveraging multi-view learning and dual-path architectures to adapt to varying slice thicknesses and overlaps.
Inspired by the success of Transformers in vision tasks, TVSRN~\cite{TVSRN} introduced a transformer-based network with an asymmetric encoder for slice interpolation.
Implicit neural representation (INR) approaches, such as ARSSR~\cite{ARSSR} and CycleINR~\cite{cycleinr}, modeled volumetric medical images as continuous functions to achieve arbitrary-scale super-resolution without relying on grid-based upsampling.
Recent developments like CuNeRF~\cite{cunerf} applied localized neural radiance fields to adaptively reconstruct fine-grained volumetric details, while I3Net~\cite{I3Net} introduced a dual-branch framework to simultaneously capture inter-slice and intra-slice dependencies.
While these methods have advanced CT slice interpolation and super-resolution, a critical limitation remains: most of them do not explicitly account for the anisotropic properties of 3D CT volumes in their architectural designs.

\subsection{Reference-based super-resolution}
Most previous works have focused on using a single reference image~\cite{CrossNet,SRNTT,yan2020towards,C2Matching}, but recently, there have been attempts to use multiple reference images which are often available for applications~\cite{LMR}.
In the medical field, there have been studies using reference image-based approaches for image-to-image translation of chest CT~\cite{chaudhary2024lungvit} and unpaired medical image enhancement~\cite{HQG-Net}.
However, existing reference-based methods generally rely on the availability and relevance of external reference images, which can lead to domain mismatch issues.
In contrast, our method exclusively leverages intra-volume axial slices as references, providing a naturally coherent context without the need for external datasets.
Although conceptually related to the Spatial Aware Filtering Module (SAFM) in LMR~\cite{LMR}, which uses a fused reference feature and learns implicit filtering weights through a nonlinear mapping, our Relevance-Adaptive Fusion (RAF) module explicitly computes per-query relevance scores across multiple intra-volume references and adaptively fuses the features.
This explicit and multi-reference design allows our framework to better exploit the diverse contextual information available within the same volume, which is particularly important for anisotropic CT interpolation tasks.

\subsection{Diffusion-Based Medical Image Synthesis}
Diffusion models have recently emerged as powerful alternatives to GAN-based approaches for medical image generation. Foundational works like DDPM~\cite{ho2020denoising} have been adapted for medical applications, with models such as MedSyn~\cite{MedSyn} introducing text-guided anatomy-aware synthesis, and MT-DDPM~\cite{pan20232d} and GH-DDM~\cite{GH-DDM} employing transformer-based or hybrid architectures to improve synthesis quality.
Additionally, diffusion models have been applied to specialized tasks, such as trabecular bone microstructure recovery~\cite{Microstructure} and CBCT-to-synthetic CT translation~\cite{CBCT-Based} for adaptive radiotherapy.
While diffusion-based methods demonstrate strong capabilities in direct image synthesis, anatomy augmentation, and artifact removal, they primarily focus on generating entire images or performing conditional translations based on external priors.
Recently, however, diffusion-based approaches have also been explored for through-plane super-resolution in brain MRI. Zhou et al.\cite{zhou2023clinical} proposed a 2D slice-wise diffusion framework that combines a coarse super-resolution network with a score-based diffusion model and plug-and-play total variation regularization, demonstrating improved 3D consistency on real clinical MRI datasets. Similarly, He et al.\cite{he2023inverseSR} introduced InverseSR, a 3D brain MRI super-resolution method based on latent diffusion models (LDMs), which leverages generative priors trained on large-scale datasets to reconstruct high-resolution volumes from sparse clinical inputs through inversion in latent space.
These methods, while effective, typically rely on generative priors or iterative inversion schemes.
In contrast, our work addresses through-plane super-resolution in anisotropic 3D CT volumes by leveraging only intra-volume axial references, ensuring structural consistency within the same patient volume and providing deterministic enhancement without relying on stochastic generation processes.

\section{Method}
We present ACVTT, a framework that interpolates CT slices with explicit cross-view texture transfer for high-resolution through-plane image reconstruction.
To effectively transfer high-frequency details from multiple in-plane images to through-plane images, we develop MRNLA based on relevance-aware adaptive fusion with non-local operations.
Fig.~\ref{fig2} shows the overall framework of the proposed ACVTT.

\subsection{Anisotropic Cross-View Texture Transfer}
Given a sparsely sampled input CT volume $V_{lr}\in \mathbb{R}^{d \times H \times W}$, the goal is to reconstruct a densely sampled CT volume $V_{hr}\in \mathbb{R}^{D\times H \times W}$ by generating intermediate slices.
Specifically, given that $V_{lr}$ is sparsely sampled by a factor of $r$ along the depth axis, we produce $r-1$ intermediate slices between every two consecutive slices, yielding a volume with total slices $D=r(d-1)+1$.
We can obtain 2D images from $V_{lr}$ in three different planes: axial, coronal, and sagittal views, denoted by $\{I_{ax}^{z}\in \mathbb{R}^{H \times W}\}_{z=1}^{d}$, $\{I_{cor}^{y}\in \mathbb{R}^{d \times W}\}_{y=1}^{H}$, and $\{I_{sag}^{x}\in \mathbb{R}^{d \times H}\}_{x=1}^{W}$, respectively. 
We first select $N$ axial slices, denoted as $\mathbf{I}_{ax} = \{{I}_{ax}^{z}\}_{z\in \mathcal{S}}$, where $\mathcal{S}$ is the set of sampled slice indices with $|\mathcal{S}|=N$, to help restore high-frequency detail in through-plane images, $I_{cor}^{y}$ and $I_{sag}^{x}$.
The coronal and sagittal view images are bilinearly upsampled to the resolution of the target high-resolution images along the depth axis, expressed as $I_{cor\uparrow}^{y}\in \mathbb{R}^{D \times W}$ and $I_{sag\uparrow}^{x}\in \mathbb{R}^{D \times H}$, respectively.
Then, we extract features for the three views through a shared encoder $E$:

\begin{equation}
F_{cor}^{y}=E(I_{cor\uparrow}^{y}), \mathbf{F}_{ax}=\{E(I_{ax}^{z})\}_{z\in\mathcal{S}}, F_{sag}^{x}=E(I_{sag\uparrow}^{x}),
\end{equation}
where $F_{cor}^{y} \in \mathbb{R}^{D \times W \times C}$ and $F_{sag}^{x} \in \mathbb{R}^{D \times H \times C}$ are the feature maps for coronal and sagittal views with the channel dimension $C$, and $\mathbf{F}_{ax} \in \mathbb{R}^{N\times H \times W \times C} $ denotes the set of axial slice features.
Here, we propose to use multiple high-resolution in-plane features $\mathbf{F}_{ax}$ as references and transfer their texture details to through-plane images to obtain enhanced through-plane features, $\tilde{F}_{cor}^{y}$ and $\tilde{F}_{sag}^{x}$:

\begin{equation}
\tilde{F}_{cor}^{y}=\mathcal{T}(F_{cor}^{y}, \mathbf{F}_{ax}), \quad \tilde{F}_{sag}^{x}=\mathcal{T}(F_{sag}^{x}, \mathbf{F}_{ax}),
\end{equation}
where $\mathcal{T}(\cdot)$ denotes a cross-view texture transfer function. 
For $\mathcal{T}(\cdot)$, we propose MRNLA, which is described in the next subsection.
The enhanced features are then passed through a shared decoder $D$ to produce reconstructed high-resolution through-plane images, which can be expressed as $\tilde{I}_{cor}^{y}=D(\tilde{F}_{cor}^{y})$ and $\tilde{I}_{sag}^{x}=D(\tilde{F}_{sag}^{x})$, respectively.
Finally, the predicted 3D CT volume for coronal and sagittal views, $\tilde{V}_{cor}=\{\tilde{I}_{cor}^{y}\}_{y=1}^{H}$ and $\tilde{V}_{sag}=\{\tilde{I}_{sag}^{x}\}_{x=1}^{W}$, can be constructed by stacking the individual through-plane enhancement results.
To obtain the final single output volume $\tilde{V}$ from the two reconstructed volumes $(\tilde{V}_{cor}, \tilde{V}_{sag})$, we adopt a residual fusion strategy that performs slice-wise averaging and residual learning, which can be written as: $\tilde{V}=\frac{1}{2}(\tilde{V}_{cor} + \tilde{V}_{sag}) + \Phi(\mathrm{concat}(\tilde{V}_{cor}, \tilde{V}_{sag})) $, where $\Phi$ represents the slice-wise convolution operations.

\begin{figure}[]
\centering
\includegraphics[width=0.8\linewidth]{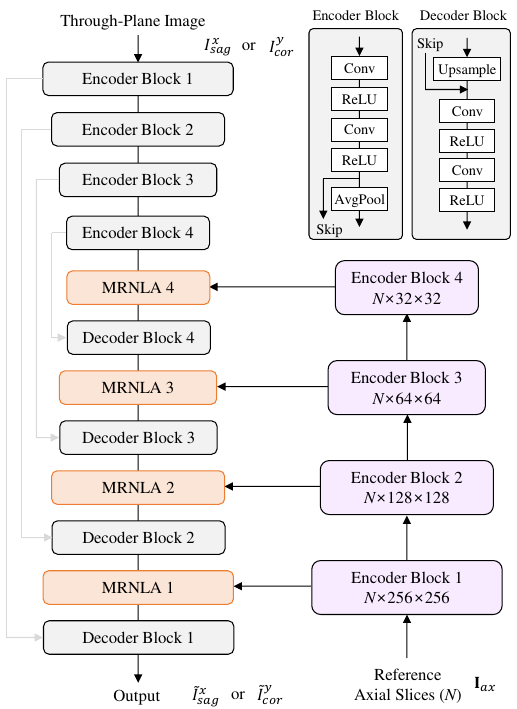}
\caption{Architecture diagram of our proposed ACVTT. The through-plane image is processed through a series of encoder blocks, while $N$ reference axial slices are also passed through corresponding encoder blocks (illustrated here for slices with a spatial resolution of 256$\times$256). Our proposed multi-reference non-local Attention (MRNLA) modules are applied to fuse features from the same level across both encoder paths, enhancing the through-plane features. The enhanced features are subsequently passed through decoder blocks, culminating in the final output.
} \label{fig4}
\end{figure}

\begin{figure*}[t]
\centering
\includegraphics[width=0.9\textwidth]{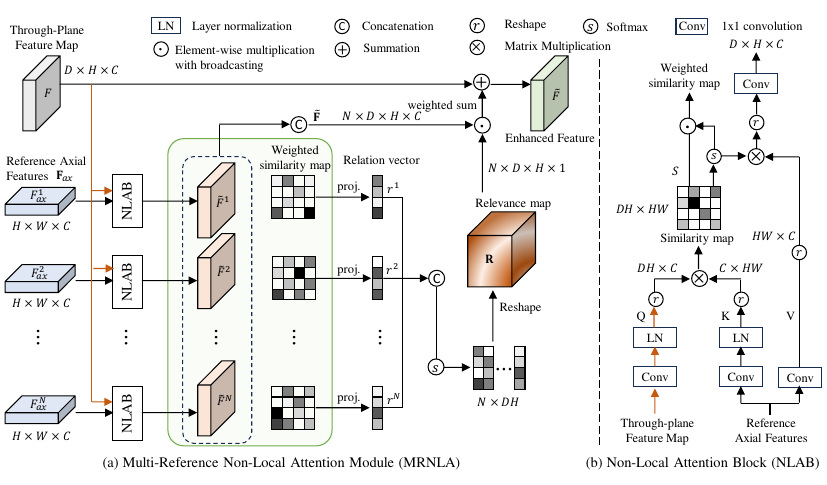}
\caption{Detailed architecture of the multi-reference non-local attention (MRNLA) module. It enhances through-plane features by adaptively transferring texture details from multiple axial slice features. The through-plane feature map interacts with each axial feature via a non-local attention block (NLAB), generating slice-wise transferred features and corresponding attention-weighted similarity maps. These similarity maps are further projected to derive relevance scores, which quantify the importance of each axial reference for specific query points. The final fused feature map is obtained through a weighted combination of transferred features, guided by relevance scores, and with a residual connection, and integrated with a residual connection to preserve original information while incorporating high-frequency details from the most relevant axial slices.
} \label{fig3}
\end{figure*}

Fig.~\ref{fig4} illustrates the detailed architecture of the proposed ACVTT framework, designed for effective feature integration between through-plane images and reference axial slices. The through-plane image is processed through a series of encoder blocks, while $N$ reference axial slices, each with a spatial resolution of 256$\times$256, are fed through parallel encoder paths. These axial slices provide high-resolution, high-frequency details, which are critical for capturing fine-grained spatial information.
The encoder and decoder blocks in the architecture are composed of two convolutional layers followed by ReLU activations~\cite{relu}, augmented by skip connections to preserve spatial details across different stages. Downsampling in the encoder is performed using average pooling, while bilinear upsampling is used in the decoder for resolution recovery. At each level of the encoder, the proposed multi-reference non-local attention (MRNLA) modules are applied, enabling cross-path feature fusion by leveraging the detailed, high-resolution information from axial slices to enhance the through-plane features. The enhanced features are progressively reconstructed through the decoder blocks, resulting in the final output. 
This architecture ensures robust feature alignment and enhancement, leading to improved performance in tasks requiring fine-grained spatial and contextual understanding.

For both through-plane and axial inputs, we use a shared encoder network $E$, which is fully trainable.
Axial reference slices are passed through this encoder during training, and the weights are jointly optimized along with the rest of the network.
This design ensures that the extracted features from axial views are well-aligned with the target features, enabling effective relevance modeling and fusion.

To generate low-resolution volumes for training, we follow two strategies depending on dataset availability.
For datasets without paired thick-thin scans, such as KiTS23~\cite{kits21} and MSD~\cite{MSD}, we synthetically downsample the high-resolution thin-slice CT volumes along the through-plane direction to simulate low-resolution inputs.
In contrast, for the RPLHR-CT~\cite{TVSRN} dataset, which contains real paired scans, we directly use the thick slice as the low-resolution input and the thin slice as the corresponding ground-truth.
This setup ensures consistency in supervision across both synthetic and real-pair training scenarios.

\subsection{Multi-Reference Non-Local Attention}
One of the key components of the proposed ACVTT is the MRNLA module that transfers texture details from multiple axial slices to the through-plane views.
Our motivation for MRNLA is to perform texture transfer adaptively according to the relevance of references to a target point in a low-resolution image.
Our MRNLA module is conceptually inspired by the non-local operation introduced in Non-local Neural Networks~\cite{nonlocal}. While building upon this foundation, MRNLA introduces two key innovations tailored to anisotropic CT slice interpolation. First, unlike the original formulation that computes non-local attention within a single image, MRNLA performs cross-plane attention, where through-plane target points serve as queries and in-plane reference points serve as keys and values. Second, MRNLA incorporates an adaptive aggregation mechanism that selectively combines information from multiple in-plane reference slices based on their relevance to the target. These extensions enable MRNLA to more effectively model long-range dependencies across anisotropic slices.
Fig.~\ref{fig3} shows the architecture of the proposed MRNLA in detail.
Given a through plane feature $F$, where $F$ denotes either $F_{cor}^{y}$ or $F_{sag}^{x}$, and multiple axial features $\mathbf{F}_{ax}=\{F_{ax}^{l}\}_{l=1}^{N}$ as input, MRNLA produces the enhanced feature map $\tilde{F}$.
Here, we explain the case of input sagittal view for convenience, \textit{i.e.,} $F\in\mathbb{R}^{D\times H\times C}$.
We first pass $F$ through a non-local attention block (NLAB) with each axial feature $F_{ax}^{l}$ to obtain slice-wise transferred feature map $\tilde{F}^{l}$ and similarity map ${S}^{l}$.
In NLAB, $F$ acts as a query $Q$, and the axial feature $F_{ax}^{l}$ as a key $K^{l}$ and a value $V^{l}$ to perform non-local operations.
Specifically, we first linearly project $F$ and $F_{ax}^{l}$ by $1\times1$ convolutions as follows: $Q=W_q(F), K^l=W_k(F_{ax}^{l}), V^l=W_v(F_{ax}^{l})$, where $W_q, W_k, W_v$ are convolutions for query, key, and value embedding, respectively.
We then apply layer normalization to query $Q$ and key $K^{l}$ and reshape the features to obtain $Q\in \mathbb{R}^{DH\times C}$, $K^{l}\in \mathbb{R}^{C\times HW}$, and $V^{l}\in \mathbb{R}^{HW\times C}$.
The similarity map $S^{l}$ and transferred feature $\tilde{F}^{l}$ are then computed as follows:

\begin{equation}
S^{l} = \frac{Q \cdot K^{l}}{\sqrt{C}},\quad \tilde{F}^{l}=W_{out}\left(\mathrm{Softmax}(S^{l})\cdot V^{l}\right),
\end{equation}
where $W_{out}$ is another $1\times1$ convolution for output projection.
Here, $S^{l}$ contains the raw similarity scores between $Q$ and $K^{l}$.
We assume that for each query point in $Q$, more helpful reference slices for through-plane high-frequency reconstruction would be different.
To quantify the relevance of references to query points, we propose to utilize $S^{l}$ as they contain similarity scores with all reference keys for each query point.
Specifically, we first project each similarity map $S^{l}\in \mathbb{R}^{DH\times HW}$ into a relation vector $r^{l}\in \mathbb{R}^{DH}$ by taking an attention-weighted average along the key axis, where the weights are obtained by applying the softmax operation over the similarity map, as follows:

\begin{equation}
r^{l} = \mathrm{Sum}_k\left(\mathrm{Softmax}(S^{l}) \odot S^{l}\right), 
\end{equation}
where $\odot$ denotes element-wise multiplication, and $\mathrm{Sum}_k$ denotes the summation along the key dimension.
In this way, $r^l$ indicates the representative similarity score of the $l^{th}$ reference slice for each point in Q.
Next, relation vectors from multiple references $\{r^{l}\}_{l=1}^{N}$ are concatenated and the softmax is applied along the first dimension to obtain a relevance map $\mathbf{R}\in \mathbb{R}^{N \times DH}$. 
Here, $\mathbf{R}$ represents the relative relevance of references to each query point.
Finally, the set of transferred features $\{\tilde{F}^{l}\}_{l=1}^{N}$ is concatenated as $\tilde{\mathbf{F}}\in \mathbb{R}^{N\times D \times H \times C}$ and then weighted averaged by the reshaped relevance map $\mathbf{R}\in \mathbb{R}^{N\times D \times H \times 1}$ to obtain the final relevance-adaptive fused feature $\tilde{F}$ with residual connection, expressed as:

\begin{equation}
\tilde{F} = \mathrm{Sum}(\mathbf{R} \odot \tilde{\mathbf{F}}) + F,
\end{equation}
where $\odot$ is performed with broadcasting and $\mathrm{Sum}$ denotes the summation along the first (reference batch) dimension.

\begin{figure}[]
\centering
\includegraphics[]{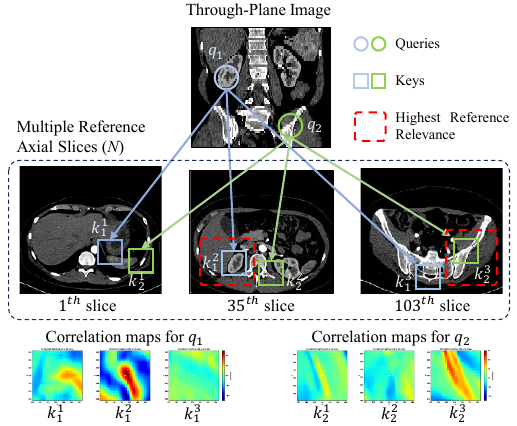}
\caption{Conceptual visualization of the MRNLA mechanism. Each query point in the through-plane image (circle) is shown alongside representative key points (rectangles) from sampled reference axial slices. The reference slice with the highest relevance to each query point is highlighted with a red bounding box. Additionally, heatmaps representing correlation maps between query points and reference features are visualized to illustrate the adaptive selection process for transferring detailed texture and contextual information.
} \label{fig5}
\end{figure}

Fig.~\ref{fig5} offers a visual representation of how the proposed MRNLA operates to dynamically enhance through-plane feature. The MRNLA mechanism adaptively selects relevant texture information from sampled reference axial slices to enhance the features of through-plane images. 
For each query point in the through-plane image, key points from multiple reference slices are analyzed. The reference slice exhibiting the highest relevance to a given query is highlighted.
This context-sensitive approach ensures that texture details are transferred in a manner tailored to the specific spatial and structural context of the query. By relying solely on intra-volume reference slices, the MRNLA module eliminates the need for external datasets, increasing the robustness and generalizability of the model. The utilization of in-plane references also guarantees that the reconstructed output remains consistent with the original imaging characteristics, producing results that are both coherent and anatomically plausible.
This design not only enhances the model's ability to reconstruct high-resolution features but also ensures efficiency and practicality in real-world applications. In summary, the figure encapsulates the novel contribution of MRNLA: a mechanism that integrates local and non-local dependencies while dynamically prioritizing information from the most relevant references to achieve superior texture transfer and feature enhancement.

Unlike standard self-attention mechanisms used in transformers, which compute intra-view token-to-token correlations within the same feature map, our MRNLA explicitly models cross-view dependencies by establishing attention between through-plane features (queries) and axial slice features (keys and values).
Furthermore, while transformer self-attention uniformly aggregates information based on similarity scores across all tokens, MRNLA introduces a relevance-adaptive fusion mechanism that selectively integrates axial references based on their contextual relevance to each query point.

Although the high-resolution axial slices only intersect with specific rows in the through-plane views, our method does not rely on direct spatial overlap between planes. Instead, it exploits the rich high-frequency patterns and contextual structures embedded in densely spaced axial slices as global priors to enhance orthogonal views. The MRNLA module computes global correlation between query points in the target plane and reference axial features, enabling the transfer of anatomically meaningful textures across views. This context-aware matching is essential due to the orientation differences between axial and through-plane slices and is in line with many reference-based super-resolution methods that perform extensive patch matching to retrieve high-frequency details. While the global correlation computation increases computational cost, this is a deliberate trade-off to overcome fundamental limitations in earlier approaches that only exploit adjacent or overlapping slices. The effectiveness of this strategy is supported by our experimental results in Section IV, where we observe consistent improvements in both reconstruction accuracy and downstream clinical task performance.

\begin{table*}[t]
\centering
\caption{Quantitative comparison on RPLHR-CT, MSD, and KiTS23 datasets. PSNR is in dB. Training time (days), inference time (s), and GPU memory usage (GB) are also reported. GPU memory is measured with batch size = 1 during training. $*$ indicates \textit{p} < 0.05 based on paired t-test.}\label{tab1}
\begin{tabular}{ c| l | c c|c c|c c | c  c  c}
\hline
\multirow{2}{*}{Methods} & \multirow{2}{*}{Ref.}  & \multicolumn{2}{|c|}{RPLHR-CT } & \multicolumn{2}{|c|}{MSD} & \multicolumn{2}{|c|}{KiTS23} & \multirow{2}{*}{Train} & \multirow{2}{*}{Infer} & \multirow{2}{*}{GPU Mem.}\\
\cline{3-8}
 & & PSNR~$(\uparrow)$ & $\mathrm{SSIM_a}$~$(\uparrow)$ & PSNR~$(\uparrow)$ & $\mathrm{SSIM_a}$~$(\uparrow)$  & PSNR~$(\uparrow)$  & $\mathrm{SSIM_a}$~$(\uparrow)$ \\
\hline
Bicubic      &  - & 34.85$^*$ & 0.9056$^*$ & 38.48$^*$ & 0.9390$^*$  &  38.13$^*$ & 0.9353$^*$ & - & - & -\\
2D MDSR ~\cite{MDSR}  & CVPR'17  & 36.47$^*$ & 0.9069$^*$  & 41.08$^*$ & 0.9605$^*$    &  40.87$^*$ & 0.9585$^*$ & 0.95 & 19.80 & 1.6\\
3D MDSR ~\cite{MDSR}    &  CVPR'17    & 37.18$^*$ & 0.9166$^*$ & 41.14$^*$ & 0.9619$^*$  &  40.25$^*$ & 0.9568$^*$ & 1.90& 33.75 & 2.3\\
2D RDN ~\cite{RDN}    & CVPR'18  & 37.80$^*$ & 0.9264$^*$ & 41.10$^*$ & 0.9604$^*$  &  40.90$^*$ & 0.9592$^*$ & 1.65 & 25.00 & 2.2 \\
3D RDN  ~\cite{RDN}   &   CVPR'18    & 36.97$^*$ & 0.9121$^*$   & 41.05$^*$ &	0.9613$^*$ &  40.87$^*$ & 0.9596$^*$ & 3.10 & 45.62 & 2.9 \\
Meta-SR ~\cite{Meta-SR}   & CVPR'19   & 37.19$^*$ & 0.9207$^*$ & 39.57$^*$ & 0.9496$^*$ &  40.74$^*$ & 0.9585$^*$ & 1.10 & 22.35 & 1.7\\
SAINT ~\cite{SAINT}    &  CVPR'20    &  37.90$^*$ & 0.9234$^*$ & 41.29$^*$ & 0.9633$^*$ & \underline{41.16}$^*$ & 0.9619$^*$ & 4.98 & 89.63 & 1.4\\
DA-VSR ~\cite{DA-VSR}   &  MICCAI'21  & 37.44$^*$ & 0.9176$^*$ & 40.10$^*$ &	0.9566$^*$  &  39.11$^*$ & 0.9529$^*$ & 2.85 & 38.10 & 2.4\\
Fang \textit{et al.}~\cite{Incremental} & CVPR'22  & 36.19$^*$ & 0.9128$^*$  & 40.51$^*$ & 0.9616$^*$ &  39.75$^*$ & 0.9510$^*$ & 1.40 & 27.65 & 2.0\\
TVSRN~\cite{TVSRN}  & MICCAI'22  & \underline{38.61}$^*$ & \underline{0.9360}$^*$ & 41.28$^*$ & 0.9663$^*$  &  40.92$^*$ & 0.9651$^*$ & 2.10 & 36.22 & 2.5 \\
SR4ZCT~\cite{sr4zct}  & MLMI'23  & 35.95$^*$ & 0.9087$^*$  & 39.64$^*$ & 0.9585$^*$    &  39.35$^*$ & 0.9527$^*$ & - & 935.0 & 1.6 \\
ARSSR~\cite{ARSSR}  & J-BHI'23  & 38.15$^*$ & 0.9317$^*$  & 40.60$^*$ & 0.9613$^*$    &  40.30$^*$ & 0.9609$^*$ & 1.62 & 28.14 & 2.1\\
InverseSR~\cite{he2023inverseSR}  & MICCAI'23  & 38.10$^*$ & 0.9310$^*$  & 40.50$^*$ & 0.9612$^*$    & 40.15$^*$ & 0.9574$^*$ & 4.50 & 980.0 & 4.2\\ 
I3Net~\cite{I3Net}  & TMI'24  & 38.40$^*$ & 0.9351$^*$  & \underline{41.34}$^*$ & \underline{0.9669}$^*$    &  41.03$^*$ & \underline{0.9656}$^*$ & 2.46 & 29.84 & 2.7\\
MFER~\cite{MFER}  & CIBM'24  & 38.25$^*$ &  0.9331$^*$  &  41.20$^*$ &  0.9640$^*$    &   40.85$^*$ &  0.9625$^*$ & 2.00& 33.50 & 2.6\\
TDAFD~\cite{TDAFD}  & J-BHI'25  &  38.35$^*$ &  0.9345$^*$  &  41.30$^*$ &  0.9645$^*$    &   41.10$^*$ &  0.9642$^*$ & 2.20 & 34.75 & 2.8\\
Zhou \textit{et al.}~\cite{zhou2023clinical}  & MLMI'25 & 38.20$^*$ & 0.9328$^*$  & 40.65$^*$ & 0.9620$^*$    & 40.35$^*$ & 0.9588$^*$ & 3.80 & 620.0 & 3.5\\ 
\textbf{ACVTT (Ours)} & -  & \textbf{39.07} & \textbf{0.9401} & \textbf{42.26} & \textbf{0.9700}  & \textbf{41.84} & \textbf{0.9688} & 1.70 & 31.00 & 2.8\\
\hline
\end{tabular}
\end{table*}

\subsection{Loss Functions}
We optimize our framework using L1 loss between the ground-truth HR volumes and the predicted volume at each view by cross-view texture transfer, which is expressed as $\mathcal{L}_{trans}=\mathcal{L}_{1}(V, \tilde{V}_{cor}) + \mathcal{L}_{1}(V, \tilde{V}_{sag})$, and another L1 loss for the fused volume $\tilde{V}$, which is written by  $\mathcal{L}_{fuse}=\mathcal{L}_{1}(V, \tilde{V})$. 
Our training process is divided into two stages: in the first stage, the network is trained to reconstruct coronal and sagittal views using a shared-weight model, with random sampling of the target view at each iteration; in the second stage, the fusion module is trained to combine the reconstructed outputs from both directions.
Thus, the overall loss is written as
$\mathcal{L} = \mathcal{L}_{trans} + \mathcal{L}_{fuse}$

\section{Experiments}

\subsection{Datasets}
We conduct experiments on three publicly available CT datasets: RPLHR-CT~\cite{TVSRN}, Medical Segmentation Decathlon challenge (MSD)~\cite{MSD}, and 2023 Kidney and Tumor Segmentation Challenge (KiTS23)~\cite{kits21}.

\subsubsection{RPLHR-CT}
The RPLHR-CT dataset is the first real-paired thin- and thick-CT benchmark for CT slice interpolation. 
It consists of paired CT scans acquired with different slice thicknesses, where scans with both thick slices (\textit{i.e.,} 5 mm) and thin slices (\textit{i.e.,} 1 mm) are included.
These pairs enable us to train and evaluate models by providing a ground truth reference (thin slices) against which interpolated results (from thick slices) can be compared.
The dataset contains 100/50/100 cases for train/val/test splits, providing a comprehensive set of data that includes a wide range of anatomical variations and imaging conditions.
Each CT scan in the dataset consists of slices with dimensions of 512×512 pixels. The number of slices varies depending on the slice thickness: thin CT scans contain between 191 and 396 slices, while thick CT scans comprise 39 to 80 slices. Both thin and thick CT scans maintain the same in-plane resolution, which ranges from 0.604 mm to 0.795 mm.

\subsubsection{MSD}
The MSD dataset contains a diverse collection of medical imaging data for different anatomical structures. 
We utilize CT scans of the liver, pancreas, hepatic vessels, spleen, and colon.
The in-plane resolution of these volumes ranges from 0.58 mm to 1.0 mm, while the through-plane resolution ranges from 0.8 mm to 8.0 mm. This indicates a significant anisotropy in the 3D CT volumes, with generally poorer through-plane resolution.
We select 240 thin CT scans with a slice thickness of 1.5 mm or less, where the volumes are then interpolated to 1mm.
We utilize 100 cases for testing, 40 for validation, and the remaining scans for training.

\subsubsection{KiTS23}
The KiTS23 dataset comprises a total of 500 CT scans who have undergone nephrectomy for kidney tumors, which exhibit a slice thickness ranging from 1.0 mm to 5.0 mm, and 0.5 mm to 1.0 mm in the in-plane resolution.
We also selected 145 thin CT scans with a slice thickness of 1.5 mm or less and resampled to 1mm.
Among them, 50 cases were used for testing, 15 for validation, and the remaining scans for training.

Both MSD and KiTS23 were originally designed for segmentation tasks and do not provide paired thick- and thin-slice CT volumes. To construct a supervised slice interpolation setting, we selected only cases with through-plane resolution $\le1.5$ mm and synthetically downsampled them by sampling axial slices with the ratio of $r$ to generate pseudo thick-slice inputs. Cases with inherently thick slice spacing (e.g., up to 8.0 mm) were excluded due to the absence of corresponding high-resolution ground truth. This strategy enables controlled evaluation while ensuring meaningful supervision.

\subsubsection{IXI}
In addition, we conduct extended experiments on the IXI\footnote{\url{http://brain-development.org/ixi-dataset}} dataset, a publicly available collection of 3D brain MR volumes. The dataset includes three imaging modalities: proton density (PD), T1-weighted, and T2-weighted scans. We use 185 T2-weighted volumes from the Hammersmith Hospital subset, acquired using a Philips 3T scanner. These are split into 150 volumes for training and 35 for testing. Most volumes have an in-plane resolution of 0.9375 mm and a through-plane resolution of 1.25 mm. Following the same protocol as in the CT experiments, we generate low-resolution input volumes by downsampling along the through-plane axis to simulate anisotropic resolution.

\subsection{Implementation Details}
All experiments are performed in Python 3.8.14 with Pytorch 1.8 on a single NVIDIA GeForce RTX 3090 GPU card with 24GB of memory.
For competing methods, the codes are adapted from publicly available implementation to our experiment setting.
For MSD and KiTS23, we pre-process the data by clipping the HU value to the range of [–1024, 3071] and then linearly scaling it to [0, 1].
We use foreground regions of the CT slices, excluding air regions, by thresholding the mean intensities of through-plane images for effective training and evaluation.
In the training stage, we randomly crop sub-volumes with the size of 256×256×256 and downsample volumes with the slice sampling ratio of $r$.
On the cropped volumes, we randomly choose through-plane images to be interpolated and randomly select $N$ axial slices for reference. 
We note that axial slices are uniformly sampled during inference.
During training, random sampling of axial reference slices encourages the model to generalize across diverse spatial configurations and increases robustness to the variation of reference selection.
In contrast, during inference, uniform sampling ensures stable coverage across the volume and consistent reconstruction performance.
For the encoder $E$ and decoder $D$, we use the standard U-Net architecture~\cite{UNet} with three downsample steps.
Our cross-view transfer is performed at each feature level of the decoder $D$.
For $\Phi$, which represents the final fused volume obtained by combining the reconstructed coronal and sagittal outputs through slice-wise convolutions, we also adopt the U-Net structure. 
We use Adam optimizer with a learning rate of $1.0\times10^{-4}$ to train our ACVTT for 20K epochs.
In our experimental setting, $r=5$ (the through-plane downsampling factor) and $N=3$ (the number of intra-volume axial reference slices, determined through ablation study).


\begin{table}[t]
\setlength{\tabcolsep}{3pt}  
\centering
\caption{View-specific SSIM scores ($\mathrm{SSIM_c}$, $\mathrm{SSIM_s}$) on RPLHR-CT, MSD, and KiTS23 datasets. $*$ indicates \textit{p} < 0.05.}
\label{tab:view_ssim_c_s}
\begin{tabular}{l|cc|cc|cc}
\hline
\multirow{2}{*}{Method} & \multicolumn{2}{c|}{RPLHR-CT} & \multicolumn{2}{c|}{MSD} & \multicolumn{2}{c}{KiTS23} \\
 & $\mathrm{SSIM_c}$ & $\mathrm{SSIM_s}$ & $\mathrm{SSIM_c}$ & $\mathrm{SSIM_s}$ & $\mathrm{SSIM_c}$ & $\mathrm{SSIM_s}$ \\
\hline
Bicubic & 0.8929$^*$ & 0.8932$^*$ & 0.9260$^*$ & 0.9265$^*$ & 0.9223$^*$ & 0.9225$^*$ \\
2D MDSR~\cite{MDSR} & 0.8940$^*$ & 0.8941$^*$ & 0.9481$^*$ & 0.9475$^*$ & 0.9461$^*$ & 0.9461$^*$ \\
3D MDSR~\cite{MDSR} & 0.9040$^*$ & 0.9039$^*$ & 0.9490$^*$ & 0.9498$^*$ & 0.9437$^*$ & 0.9443$^*$ \\
2D RDN~\cite{RDN} & 0.9137$^*$ & 0.9138$^*$ & 0.9480$^*$ & 0.9482$^*$ & 0.9462$^*$ & 0.9461$^*$ \\
3D RDN~\cite{RDN} & 0.8995$^*$ & 0.8990$^*$ & 0.9481$^*$ & 0.9484$^*$ & 0.9469$^*$ & 0.9467$^*$ \\
Meta-SR~\cite{Meta-SR} & 0.9081$^*$ & 0.9085$^*$ & 0.9371$^*$ & 0.9369$^*$ & 0.9460$^*$ & 0.9457$^*$ \\
SAINT ~\cite{SAINT}& 0.9108$^*$ & 0.9112$^*$ & 0.9503$^*$ & 0.9503$^*$ & 0.9489$^*$ & 0.9494$^*$ \\
DA-VSR ~\cite{DA-VSR}& 0.9053$^*$ & 0.9050$^*$ & 0.9436$^*$ & 0.9439$^*$ & 0.9402$^*$ & 0.9400$^*$ \\
Fang et al. ~\cite{Incremental} & 0.9001$^*$ & 0.8997$^*$ & 0.9486$^*$ & 0.9489$^*$ & 0.9384$^*$ & 0.9383 \\
TVSRN ~\cite{TVSRN}& 0.9232$^*$ & 0.9238$^*$ & 0.9531$^*$ & 0.9539$^*$ & 0.9520$^*$ & 0.9524$^*$ \\
SR4ZCT ~\cite{sr4zct}& 0.8961$^*$ & 0.8962$^*$ & 0.9461$^*$ & 0.9456$^*$ & 0.9404$^*$ & 0.9401$^*$ \\
ARSSR ~\cite{ARSSR}& 0.9192$^*$ & 0.9186$^*$ & 0.9489$^*$ & 0.9483$^*$ & 0.9480$^*$ & 0.9481$^*$ \\
InverseSR ~\cite{he2023inverseSR} & 0.9201$^*$ & 0.9198$^*$ & 0.9503$^*$ & 0.9500$^*$ & 0.9491$^*$ & 0.9493$^*$ \\
I3Net ~\cite{I3Net}& 0.9228$^*$ & 0.9229$^*$ & 0.9538$^*$ & 0.9541$^*$ & 0.9533$^*$ & 0.9534$^*$ \\
MFER ~\cite{MFER} & 0.9209$^*$ & 0.9206$^*$ & 0.9520$^*$ & 0.9518$^*$ & 0.9501$^*$ & 0.9502$^*$ \\
TDAFD ~\cite{TDAFD} & 0.9222$^*$ & 0.9218$^*$ & 0.9532$^*$ & 0.9529$^*$ & 0.9514$^*$ & 0.9513$^*$ \\
Zhou \textit{et al.} ~\cite{zhou2023clinical} & 0.9210$^*$ & 0.9207$^*$ & 0.9510$^*$ & 0.9507$^*$ & 0.9501$^*$ & 0.9502$^*$ \\
\textbf{ACVTT (Ours)} & \textbf{0.9276} & \textbf{0.9272} & \textbf{0.9576} & \textbf{0.9574} & \textbf{0.9563} & \textbf{0.9562} \\
\hline
\end{tabular}
\end{table}

\subsection{Comparison with State-of-the-Art Methods}
We compare our proposed ACVTT with existing state-of-the-art CT slice interpolation methods, including SAINT~\cite{SAINT}, DA-VSR~\cite{DA-VSR}, Incremental Distillation~\cite{Incremental}, TVSRN~\cite{TVSRN}, ARSSR~\cite{ARSSR}, and I3Net~\cite{I3Net}, across three datasets: RPLHR-CT, MSD, and KiTS23. Additionally, we evaluate our method against several image super-resolution algorithms (e.g., MDSR~\cite{MDSR}, RDN~\cite{RDN}, Meta-SR~\cite{Meta-SR}) and their 3D variants (3D MDSR and 3D RDN). To incorporate the latest developments in 3D reconstruction, we additionally benchmarked our model against two recent 3D SR methods — MFER~\cite{MFER} and TDAFD ~\cite{TDAFD}. 
To ensure a comprehensive evaluation that reflects recent diffusion-based advancements in slice interpolation, we also compare our method with two state-of-the-art diffusion models: the slice-wise score-based refinement framework proposed by Zhou et al.\cite{zhou2023clinical} and the latent diffusion-based reconstruction method InverseSR\cite{he2023inverseSR}.
The quality of the reconstructed CT slices is assessed using two widely-used metrics: peak signal-to-noise ratio (PSNR) and structural similarity index (SSIM~\cite{SSIM})

\begin{table}[t]
\setlength{\tabcolsep}{4pt}  
\centering
\caption{Quantitative comparison on KiTS23 datasets under multiple through-plane upscaling factors ($\times$2 to $\times$4). PSNR is in dB. $*$ indicates \textit{p} < 0.05.}
\label{tab:kits_multi}
\begin{tabular}{lcccccc}
\hline
\multirow{2}{*}{Method} & \multicolumn{2}{c}{\textbf{$\times$2}} & \multicolumn{2}{c}{\textbf{$\times$3}} & \multicolumn{2}{c}{\textbf{$\times$4}} \\
\cline{2-7}
 & PSNR & SSIM & PSNR & SSIM & PSNR & SSIM  \\
\hline
Bicubic~         & 47.19$^*$ & 0.9847$^*$ & 43.75$^*$ & 0.9769$^*$ & 39.75$^*$ & 0.9522$^*$ \\
Meta-SR~\cite{Meta-SR}        & 49.83$^*$ & 0.9916 & 46.66$^*$ & 0.9755$^*$ & 43.83$^*$ & 0.9732$^*$ \\
SAINT~\cite{SAINT}           & 50.09$^*$ & 0.9759$^*$ & 47.03$^*$ & 0.9801$^*$ & 44.23$^*$ & 0.9703$^*$ \\
DA-VSR~\cite{DA-VSR}          & 49.02$^*$ & 0.9776$^*$ & 46.17$^*$ & 0.9700$^*$ & 43.89$^*$ & 0.9696$^*$ \\
Fang et al.~\cite{Incremental}     & 49.00$^*$ & 0.9741$^*$ & 46.27$^*$ & 0.9824 & 43.08$^*$ & 0.9720$^*$ \\
TVSRN~\cite{TVSRN}           & 49.91$^*$ & 0.9872$^*$ & 46.91$^*$ & 0.9757$^*$ & 43.90$^*$ & 0.9654$^*$ \\
ARSSR~\cite{ARSSR}           & 49.65$^*$ & 0.9889$^*$ & 46.44$^*$ & 0.9745$^*$ & 43.65$^*$ & 0.9649$^*$ \\
InverseSR~\cite{he2023inverseSR}          & 49.60$^*$ & 0.9885$^*$ & 46.55$^*$ & 0.9748$^*$ & 43.75$^*$ & 0.9669$^*$ \\
I3Net~\cite{I3Net}           & 50.14 & 0.9913 & 47.07$^*$ & 0.9811$^*$ & 44.13$^*$ & 0.9706$^*$ \\
MFER~\cite{MFER} & 49.72$^*$ & 0.9893$^*$ & 46.73$^*$ & 0.9757$^*$ & 43.83$^*$ & 0.9672$^*$ \\
TDAFD~\cite{TDAFD} & 49.79$^*$ & 0.9898$^*$ & 46.88$^*$ & 0.9764$^*$ & 43.91$^*$ & 0.9686$^*$ \\
Zhou \textit{et al.}~\cite{zhou2023clinical} & 49.68$^*$ & 0.9890$^*$ & 46.70$^*$ & 0.9755$^*$ & 43.82$^*$ & 0.9675$^*$ \\

\textbf{ACVTT (Ours)} & \textbf{50.23} & \textbf{0.9922} & \textbf{47.35} & \textbf{0.9825} & \textbf{44.69} & \textbf{0.9742} \\
\hline
\end{tabular}
\end{table}

\begin{table}[t]
\setlength{\tabcolsep}{4pt}  
\centering
\caption{Quantitative comparison of reference-based methods on RPLHR-CT, MSD, and KiTS23 datasets. PSNR is in dB. $*$ indicates \textit{p} < 0.05.}
\label{tab:ref_based_methods}
\begin{tabular}{l|cc|cc|cc}
\hline
\multirow{2}{*}{Method} & \multicolumn{2}{c|}{RPLHR-CT} & \multicolumn{2}{c|}{MSD} & \multicolumn{2}{c}{KiTS23} \\
 & PSNR & $\mathrm{SSIM_a}$ & PSNR & $\mathrm{SSIM_a}$ & PSNR & $\mathrm{SSIM_a}$ \\
\hline
Bicubic & 34.8$^*$5 & 0.9056$^*$ & 38.48$^*$ & 0.9390$^*$ & 38.13$^*$ & 0.9353$^*$ \\
CrossNet~\cite{CrossNet} & 37.34$^*$ & 0.9268$^*$ & 40.92$^*$ & 0.9608$^*$ & 40.49$^*$ & 0.9548$^*$ \\
SRNTT~\cite{SRNTT} & 37.40$^*$ & 0.9273$^*$ & 40.97$^*$ & 0.9612$^*$ & 40.54$^*$ & 0.9552$^*$ \\
CIMR~\cite{yan2020towards} & 37.45$^*$ & 0.9277$^*$ & 41.02$^*$ & 0.9616$^*$ & 40.59$^*$ & 0.9556$^*$ \\
C2Matching~\cite{C2Matching} & 37.52$^*$ & 0.9283$^*$ & 41.10$^*$ & 0.9622$^*$ & 40.67$^*$ & 0.9562$^*$ \\
LMR~\cite{LMR} & 37.72$^*$ & 0.9310$^*$ & 41.30$^*$ & 0.9642$^*$ & 40.83$^*$ & 0.9577$^*$ \\
\textbf{ACVTT (Ours)} & \textbf{39.07} & \textbf{0.9401} & \textbf{42.26} & \textbf{0.9700} & \textbf{41.84} & \textbf{0.9688} \\
\hline
\end{tabular}
\end{table}

\begin{figure}[t]
\centering
\includegraphics[width=\columnwidth]{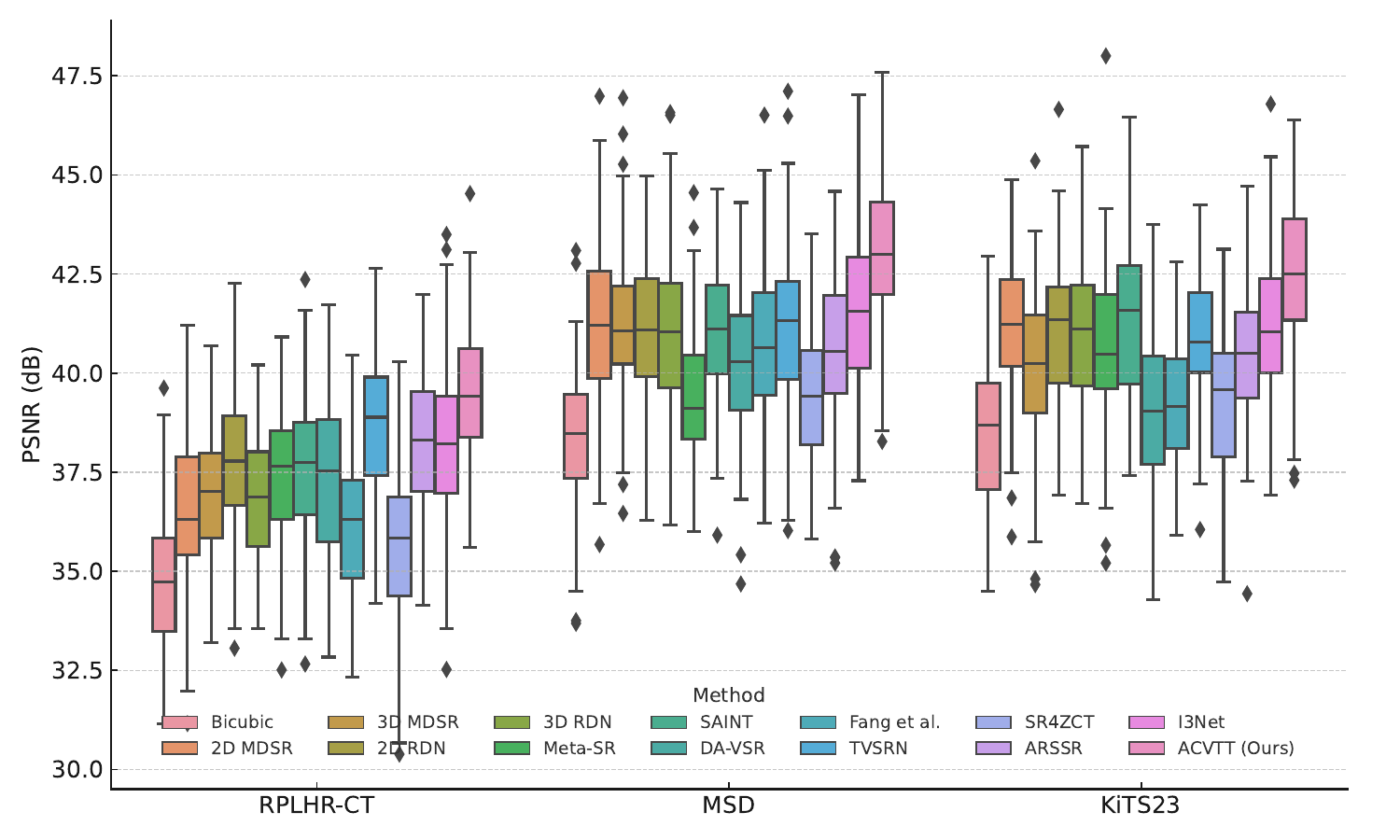}
\caption{Box plot of PSNR distributions on RPLHR-CT, MSD, and KiTS23 datasets.} \label{fig:psnr_boxplot}
\end{figure}

\begin{figure*}[!t]
\centering
\includegraphics[width=\textwidth]{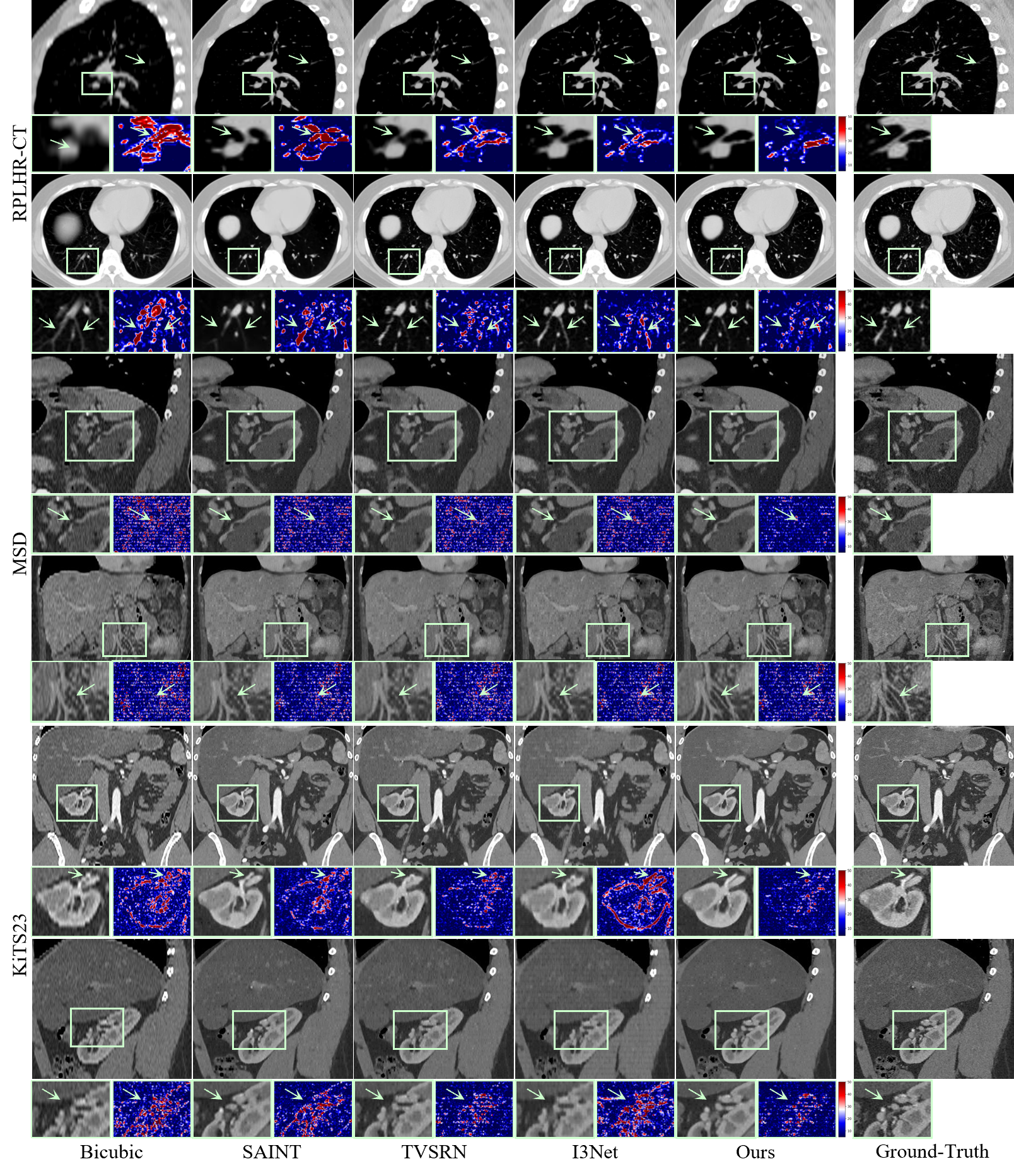}
\caption{Visual comparison between our ACVTT and state-of-the-art methods. Difference maps between the ground-truth and the reconstructed images are provided to explicitly visualize the differences, where brighter areas indicate larger differences.
Rows 1 to 6 correspond to the following view orientations:
sagittal, axial, sagittal, coronal, coronal, and sagittal, respectively.} \label{fig_4}
\end{figure*}

\begin{table*}[t]
\centering
\caption{Ablation on the relevance-adaptive fusion approach and number of sampled reference axial slices $N$ on RPLHR-CT, MSD, and KiTS23. PSNR is in dB. Inference time (sec) is reported for each $N$.}\label{tab2}
\begin{tabular}{c|c|c c|c c|c c | c}
\hline
\multicolumn{2}{c|}{Components }   & \multicolumn{2}{|c|}{RPLHR-CT } & \multicolumn{2}{|c|}{MSD} & \multicolumn{2}{|c|}{KiTS23} & \multirow{2}{*}{Time~(s)~$(\downarrow)$} \\
\cline{1-8}
$N$ & Relevance-adaptive fusion & PSNR~$(\uparrow)$ & SSIM~$(\uparrow)$ & PSNR~$(\uparrow)$ & SSIM~$(\uparrow)$  & PSNR~$(\uparrow)$  & SSIM~$(\uparrow)$ \\
\hline
0 (baseline) & - & 38.16 & 0.9321 & 41.13 & 0.9619 & 40.94 & 0.9602 & 11.19 \\
1 & - & 38.73 & 0.9375 & 41.74 & 0.9647 & 41.39 & 0.9648 & 18.47 \\
\hline
\multirow{2}{*}{2} & \xmark & 38.72 & 0.9372 & 41.78 & 0.9648 & 41.38 & 0.9642 & \multirow{2}{*}{25.17}\\
& \checkmark & 38.96 & 0.9392 & 42.13 & 0.9685  & 41.74 & 0.9676 \\
\hline
\multirow{2}{*}{3} & \xmark & 38.70 & 0.9369 & 41.75 & 0.9645 & 41.36 & 0.9642 & \multirow{2}{*}{31.00}\\
 & \checkmark & 39.07 & 0.9401 & 42.26 & 0.9700  & \textbf{41.84} & \textbf{0.9688} \\
\hline
 \multirow{2}{*}{4} & \xmark & 38.69 & 0.9368 & 41.74 & 0.9643 & 41.35 & 0.9640 & \multirow{2}{*}{36.72} \\
 & \checkmark & 39.05 & 0.9402 & \textbf{42.27} & \textbf{0.9703} & \textbf{41.84} & 0.9687 \\
\hline
\multirow{2}{*}{5} & \xmark & 38.67 & 0.9366 & 41.70 & 0.9640 & 41.33 & 0.9638 & \multirow{2}{*}{42.67}\\
 & \checkmark & \textbf{39.08} & \textbf{0.9405} & \textbf{42.27} & 0.9701 & 41.83 & 0.9685\\
\hline
\end{tabular}
\end{table*}

\begin{figure}[]
\centering
\includegraphics[width=\linewidth]{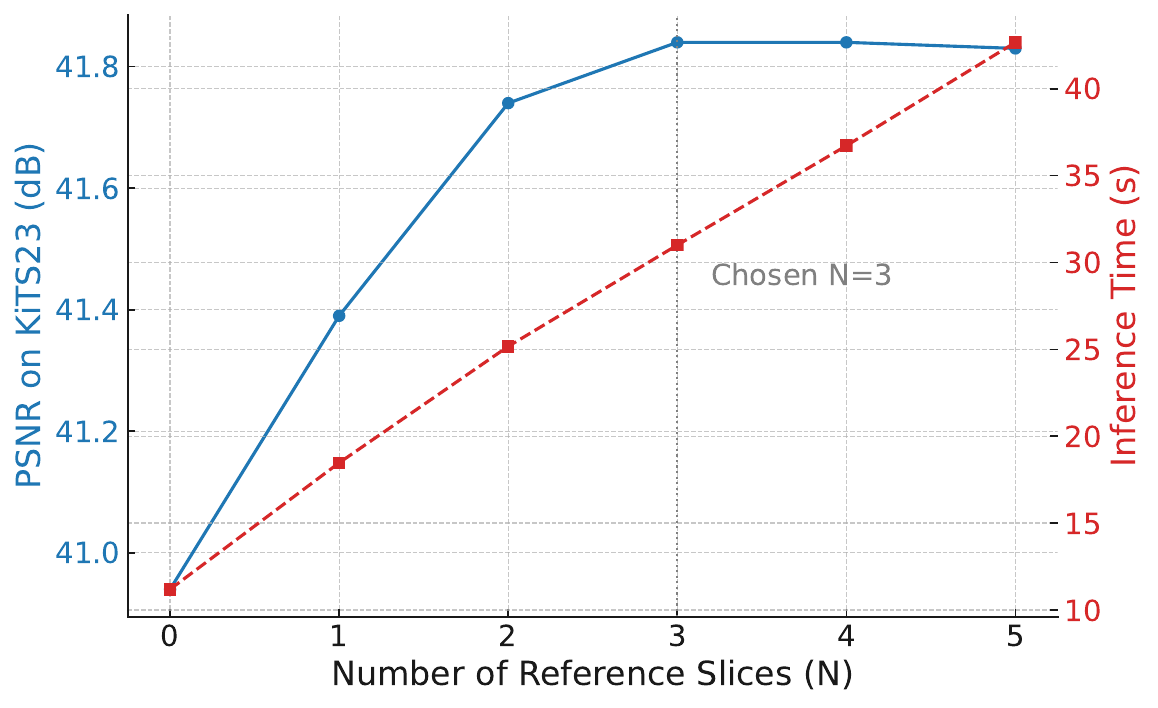}
\caption{Trade-off between reconstruction accuracy and inference time with varying numbers of reference axial slices \(N\) on the KiTS23 dataset.
} \label{fig:trade}
\end{figure}

\subsubsection{Quantitative Comparison.}
Table~\ref{tab1} presents a detailed quantitative comparison of our ACVTT with state-of-the-art methods on the RPLHR-CT, MSD, and KiTS23 datasets. The results clearly demonstrate the superiority of ACVTT across all datasets, where it consistently achieves the highest PSNR and SSIM values, outperforming both traditional image super-resolution methods and specialized CT slice interpolation approaches.
Here, the reported SSIM values correspond to the axial view ($\mathrm{SSIM_a}$). For completeness, the SSIM values for the coronal ($\mathrm{SSIM_c}$) and sagittal ($\mathrm{SSIM_s}$) views are additionally reported in Table~\ref{tab:view_ssim_c_s}.
On the RPLHR-CT dataset, ACVTT achieves a PSNR of 39.07 dB and an SSIM of 0.9401, marking a significant improvement over TVSRN, which records a PSNR of 38.61 dB and SSIM of 0.9360. These values for TVSRN are taken directly from its original paper, ensuring a fair and consistent comparison. This performance gain is particularly notable given the real-paired nature of the dataset, which reflects clinical scenarios with complex variability. The ability of ACVTT to deliver substantial improvements in such conditions highlights its practical relevance for real-world applications.
On the MSD dataset, which contains diverse anatomical structures and challenges associated with different imaging protocols, ACVTT achieves an outstanding PSNR of 42.26 dB and SSIM of 0.9700. This surpasses the second-best method, I3Net, which records a PSNR of 41.34 dB and SSIM of 0.9669. The significant margin of improvement underscores the versatility and robustness of ACVTT in handling datasets with high heterogeneity.
Similarly, on the KiTS23 dataset, which is characterized by highly detailed and irregular structures, ACVTT achieves the best performance with a PSNR of 41.84 dB and SSIM of 0.9688, compared to I3Net’s PSNR of 41.03 dB and SSIM of 0.9656. The consistent improvement across these datasets highlights the generalizability of ACVTT and its capacity to adapt to varying levels of detail and noise inherent in CT data. To assess the reliability of the observed performance improvements, we conducted paired t-tests between our method and baseline approaches across all datasets. The results indicate that the gains in PSNR and SSIM are statistically significant in most cases, with p$<$0.05.
Table~\ref{tab:view_ssim_c_s} presents the view-specific SSIM scores, reporting the coronal and sagittal SSIMs in addition to the axial values in Table~\ref{tab1}. Our method shows consistently high performance across all views, exhibiting similar trends to the axial results.

While recent 3D reconstruction methods demonstrate strong performance on isotropic 3D data such as high-resolution MRI, our experiments show that their effectiveness is limited in anisotropic CT slice interpolation. This is likely due to the mismatch between the assumptions of 3D networks and the characteristics of sparsely sampled CT volumes, where voxel spacing varies significantly across axes. Applying 3D operations without explicitly addressing such anisotropy tends to result in suboptimal performance.
This limitation stems from the inherent sampling characteristics of CT data: the in-plane resolution (typically axial) is much finer than the through-plane resolution, often resulting in large spacing gaps between slices. Recognizing this structural imbalance, we designed our approach to treat in-plane and through-plane features separately, rather than relying on isotropy-assuming volumetric operations. By decoupling the two and modeling their interactions explicitly, our method is better suited to capture the directional context necessary for accurate slice interpolation in anisotropic volumes.

To provide a more comprehensive comparison, we report the training time (in days), inference time (in seconds per full CT volume), and GPU memory usage (in GB) for all baseline and proposed methods in Table~\ref{tab1}. All metrics were measured on an NVIDIA RTX 3090 GPU under consistent software conditions, and GPU memory usage was recorded with a batch size of 1.
Despite processing multiple reference slices, our proposed ACVTT model maintains moderate computational demands, requiring 1.70 days for training, 31.00 seconds for inference, and 2.8 GB of memory. Compared to other state-of-the-art models, ACVTT achieves superior reconstruction quality while maintaining a favorable balance between accuracy and efficiency.

The reported inference times for all methods, including baselines, were measured under our unified experimental environment using the RPLHR-CT dataset, ensuring consistency and fairness. In contrast, some baseline methods such as TDAFD originally reported inference times under different conditions—typically for 2$\times$ upsampling at 256$\times$256 resolution. These are not directly comparable to our setting, which involves 5$\times$ super-resolution at 512$\times$512 resolution. Therefore, for fair comparison, we report the inference times as re-measured in our setting rather than relying on numbers reported in their original papers.

Fig.~\ref{fig:psnr_boxplot} visualizes the PSNR distribution across test samples for each method. The results demonstrate that our method (ACVTT) consistently yields the highest PSNR values and narrow variability, highlighting its robustness and superior reconstruction quality.

To further evaluate the robustness of our method, we conducted experiments using different through-plane upscaling factors ($\times$2, $\times$3, and $\times$4) on the KiTS23 dataset. The corresponding results are summarized in Table~\ref{tab:kits_multi}, showing that our method consistently achieves strong performance across varying levels of anisotropy.
Notably, ACVTT outperforms other strong methods, including SAINT, ARSSR, and I3Net. The key to ACVTT's success lies in its innovative Multi-Reference Non-Local Attention (MRNLA) mechanism, which plays a pivotal role in adaptively fusing features from multiple reference axial slices. By leveraging high-frequency textures and fine-grained details captured in the reference slices, ACVTT significantly enhances the through-plane features, enabling more accurate reconstructions. The result is not only quantitatively superior metrics but also visually coherent reconstructions that retain structural integrity and realistic textural details.
These results collectively demonstrate the strength of our cross-view texture transfer framework in reconstructing high-resolution, densely-sampled 3D CT volumes. The consistent improvement across diverse datasets and metrics underscores the adaptability and effectiveness of ACVTT, establishing it as a robust and clinically relevant solution for CT slice interpolation tasks.

We additionally benchmarked our method against several reference-based super-resolution models originally developed for 2D natural images, including CrossNet~\cite{CrossNet}, SRNTT~\cite{SRNTT}, CIMR~\cite{yan2020towards}, C2Matching~\cite{C2Matching}, and LMR~\cite{LMR}.
To ensure a fair comparison, we adapted their official implementations to our 3D CT slice interpolation setting by modifying the input structure and retraining on our medical datasets.
However, as shown in Table~\ref{tab:ref_based_methods}, these methods showed only limited performance gains in this setting.
This suggests that, although these methods are effective in the 2D natural image domain, their direct application to 3D CT image reconstruction tasks remains suboptimal.
The limited performance gain indicates a significant domain gap between natural images and medical images, which likely hinders the effectiveness of these approaches in clinical settings.


\begin{table*}[t]
\centering
\caption{Quantitative comparison on the IXI MRI dataset (T2-weighted) for slice interpolation under multiple through-plane upscaling factors ($\times$2 to $\times$5). $*$ indicates \textit{p} < 0.05.
PSNR is in dB.}
\label{tab:ixi_results}
\begin{tabular}{lcccccccc}
\hline
\multirow{2}{*}{Method} & \multicolumn{2}{c}{\textbf{$\times$2}} & \multicolumn{2}{c}{\textbf{$\times$3}} & \multicolumn{2}{c}{\textbf{$\times$4}} & \multicolumn{2}{c}{\textbf{$\times$5}} \\
\cline{2-9}
 & PSNR (↑) & SSIM (↑) & PSNR (↑) & SSIM (↑) & PSNR (↑) & SSIM (↑) & PSNR (↑) & SSIM (↑) \\
\hline
Bicubic         & 47.40$^*$ & 0.9820$^*$ & 43.80$^*$ & 0.9650$^*$ & 40.01$^*$ & 0.9450$^*$ & 36.24$^*$ & 0.9207$^*$ \\
Meta-SR~\cite{Meta-SR}         & 49.45$^*$ & 0.9940 & 46.70$^*$ & 0.9868$^*$ & 43.95$^*$ & 0.9796$^*$ & 41.20$^*$ & 0.9724$^*$ \\
SAINT~\cite{SAINT}             & 49.92$^*$ & 0.9942 & 47.05$^*$ & 0.9869$^*$ & 44.19$^*$ & 0.9797$^*$ & 41.32$^*$ & 0.9725$^*$ \\
DA-VSR~\cite{DA-VSR}          & 48.90$^*$ & 0.9940 & 46.39$^*$ & 0.9866$^*$ & 43.88$^*$ & 0.9792$^*$ & 41.27$^*$ & 0.9718$^*$ \\
Fang et al.~\cite{Incremental} & 49.35$^*$ & 0.9930$^*$ & 46.36$^*$ & 0.9862$^*$ & 43.37$^*$ & 0.9794$^*$ & 40.38$^*$ & 0.9726$^*$ \\
TVSRN~\cite{TVSRN}           & 49.88$^*$ & 0.9940 & 47.10$^*$ & 0.9869$^*$ & 44.00$^*$ & 0.9795$^*$ & 41.25$^*$ & 0.9722$^*$ \\
ARSSR~\cite{ARSSR}           & 49.78$^*$ & 0.9938 & 46.95$^*$ & 0.9865$^*$ & 43.90$^*$ & 0.9791$^*$ & 41.15$^*$ & 0.9719$^*$ \\
InverseSR~\cite{he2023inverseSR}   & 49.75$^*$ & 0.9936$^*$ & 46.85$^*$ & 0.9861$^*$ & 43.89$^*$ & 0.9789$^*$ & 41.18$^*$ & 0.9716$^*$ \\
I3Net~\cite{I3Net}           & 50.10 & 0.9941 & 47.20$^*$ & 0.9870$^*$ & 44.10$^*$ & 0.9793$^*$ & 41.39$^*$ & 0.9720$^*$ \\
MFER~\cite{MFER} & 49.70$^*$ & 0.9939 & 46.82$^*$ & 0.9864$^*$ & 43.92$^*$ & 0.9794$^*$ & 41.29$^*$ & 0.9721$^*$ \\
TDAFD~\cite{TDAFD} & 49.81$^*$ & 0.9940 & 46.98$^*$ & 0.9867$^*$ & 44.02$^*$ & 0.9798$^*$ & 41.35$^*$ & 0.9723$^*$ \\
Zhou \textit{et al.}~\cite{zhou2023clinical} & 49.79$^*$ & 0.9938$^*$ & 46.92$^*$ & 0.9863$^*$ & 43.95$^*$ & 0.9792$^*$ & 41.24$^*$ & 0.9720$^*$ \\
\textbf{ACVTT (Ours)} & \textbf{50.23} & \textbf{0.9946} & \textbf{47.55} & \textbf{0.9896} & \textbf{44.67} & \textbf{0.9836} & \textbf{42.07} & \textbf{0.9786} \\

\hline
\end{tabular}
\end{table*}

\subsubsection{Qualitative Comparison.}
We present a qualitative visual comparison of the reconstructed CT images produced by various methods, including Bicubic, SAINT, TVSRN, I3Net, and the proposed ACVTT (ours), against the ground-truth images. The results, provided in Fig.~\ref{fig_4}, cover reconstructions from the RPLHR-CT, MSD, and KiTS23 datasets, encompassing axial, coronal, and sagittal views to comprehensively evaluate the performance of each method across different perspectives.
The visual comparisons highlight the superior ability of ACVTT to generate high-resolution reconstructions that closely resemble the ground-truth images, capturing finer details and structural intricacies with exceptional clarity. In contrast, alternative methods often exhibit blurring, artifacts, or loss of high-frequency details, particularly in challenging regions with complex textures or subtle anatomical variations. 
Notably, the second row from the top shows the results for axial direction 2D slices for the RPLHR-CT dataset, where the original thick axial slices (5 mm) and target thin slices (1 mm) differ significantly. The reconstructed axial images produced by our method closely resemble the target thin CT sections.
This transformation can be beneficial for diagnostic tasks where high-resolution thin-section CT images are required.
The enhanced fidelity and precision of ACVTT’s results across multiple views and datasets underline its robustness and effectiveness in addressing the challenges of CT slice reconstruction while preserving clinically significant features.

\subsection{Ablation Study}
We conduct a comprehensive ablation study to evaluate the contribution of key components in our framework: 
(i) the number of axial reference slices ($N$),
(ii) the relevance-adaptive fusion mechanism, and
(iii) the inclusion of the MRNLA module.
Table~\ref{tab2} summarizes these controlled experiments across three datasets: RPLHR-CT, MSD, and KiTS23.

We begin with a baseline configuration ($N=0$), where no reference slices are used. As $N$ increases from 1 to 3, performance improves consistently across all datasets. For instance, on RPLHR-CT, PSNR rises from 38.73 at $N=1$ to 39.07 at $N=3$, showing the benefit of incorporating structural information from multiple in-plane references. This trend confirms that axial slices contain rich high-frequency details that effectively support through-plane reconstruction.

The relevance-adaptive fusion mechanism further enhances this process by selectively integrating the most informative features across reference slices. When enabled at $N=2$, it yields a 0.24 dB improvement in PSNR compared to a naive setup without it, highlighting its effectiveness in aggregating multi-reference textures.

To investigate the scalability of our method, we extend $N$ to 4 and 5. While minor performance gains are still observed (e.g., PSNR reaches 39.09dB at $N=5$), the improvements become marginal, suggesting a saturation point due to increasing redundancy. In contrast, inference time grows approximately linearly with $N$—from 11.19 seconds at $N=0$ to 42.67 seconds at $N=5$ per volume—introducing significant computational overhead. Based on this trade-off, we adopt $N=3$ as the default configuration, striking a balance between reconstruction quality and efficiency (Fig. ~\ref{fig:trade}).
For latency-sensitive applications, setting $N=2$ may be a practical alternative with minimal performance degradation.

Additionally, we compare the proposed relevance-adaptive fusion against a naive averaging strategy, where reference features are simply averaged:
$\tilde{F} = \mathrm{Avg}(\tilde{\mathbf{F}}) + F$.
Under this setting, performance stagnates beyond $N=1$, indicating that uniform aggregation fails to preserve the unique contributions of individual references. This emphasizes the necessity of relevance-aware feature selection.

Finally, removing the MRNLA module entirely leads to substantial degradation in both PSNR and SSIM. Since axial and coronal/sagittal views are not spatially aligned, the NLAB-based semantic matching within MRNLA is essential for accurate cross-view transfer. The combination of relevance-aware fusion and global matching enables principled integration of unaligned references—something not achievable through naive designs. These findings demonstrate that MRNLA is a critical component in realizing the full potential of reference-guided reconstruction.

\begin{table}[t]
\centering
\caption{Segmentation performance on KiTS23 to validate effectiveness on clinically relevant downstream tasks.}\label{tab3}
\begin{tabular}{c|c|c | c|c }
\hline
\multirow{2}{*}{Methods} & \multicolumn{2}{c|}{Kidney } & \multicolumn{2}{c}{Tumor }  \\
\cline{2-5}
 & Dice~$(\uparrow)$ & NSD~$(\uparrow)$ & Dice~$(\uparrow)$ & NSD~$(\uparrow)$ \\
\hline
Bicubic  & 0.9329 & 0.8409 & 0.6274 & 0.4313 \\
SAINT ~\cite{SAINT}  & 0.9346 & 0.8324 & 0.6314 & 0.4315 \\
TVSRN ~\cite{TVSRN}   & 0.9376 & 0.8429 & 0.6321 & 0.4346 \\
ARSSR ~\cite{ARSSR}  & 0.9385 & 0.8498 & 0.6364 & 0.4377 \\
I3Net ~\cite{I3Net}   & 0.9405 & 0.8554 & 0.6430 & 0.4417 \\
ACVTT (Ours)    & \textbf{0.9438} & \textbf{0.8763} & \textbf{0.6584} & \textbf{0.4559} \\
\hline
\end{tabular}
\end{table}

\subsection{Additional Evaluation on MRI Data}
To assess the generalizability of our framework beyond CT, we conducted additional experiments on the IXI MRI dataset using T2-weighted images. Although ACVTT was originally developed for CT slice interpolation, its core principle—leveraging high-resolution intra-volume references for cross-view texture transfer—is modality-agnostic and naturally extends to other 3D medical imaging modalities with anisotropic resolution.
We simulated anisotropic MRI volumes by synthetically downsampling the through-plane axis with scaling factors of $\times$2, $\times$3, $\times$4, and $\times$5, using the original high-resolution volumes as ground truth. Our method was evaluated alongside several state-of-the-art approaches under consistent training and testing conditions.
As shown in Table~\ref{tab:ixi_results}, ACVTT consistently outperformed all competing methods across all upscaling factors. The performance gap became more pronounced as the scaling factor increased, demonstrating the robustness of ACVTT in more challenging interpolation scenarios. SSIM values followed a similar trend, further validating the cross-modality generalizability and effectiveness of our approach.

\subsection{Downstream Task}
We further evaluated the clinical relevance of our method by testing its performance on a downstream tumor segmentation task. While our approach demonstrated PSNR improvements of 0.5 to 1.0 dB over state-of-the-art methods, this additional evaluation highlights its potential in capturing clinically significant features, such as tumor shape and size. For 3D segmentation, we utilized the 3D-UNet~\cite{3DUNet} model from the nnUNet~\cite{nnunet} package. Using a segmentation model trained for 1000 epochs on the training set, we measured segmentation performance on the test set with Dice scores and normalized surface distance (NSD) for both tumor and kidney segmentation across various methods, including Bicubic, SAINT, TVSRN, ARSSR, I3Net, and our proposed ACVTT. For tumor segmentation, the Dice scores were 0.6274, 0.6314, 0.6321, 0.6364, 0.6430, and 0.6584, respectively, and for kidney segmentation, the scores were 0.9329, 0.9346, 0.9376, 0.9385, 0.9405, and 0.9438. Similarly, ACVTT outperformed other methods in NSD metrics, further demonstrating its robustness in accurately capturing the shape and size of clinically significant structures. These results indicate that our ACVTT method not only achieves improved image quality but also holds clinical significance by enhancing segmentation performance.

\section{Limitations and Future Work}
In this study, we trained separate models for each through-plane upsampling factor (2$\times$, 3$\times$, 4$\times$, and 5$\times$), all using the same unified architecture. While the current implementation assumes a fixed slice thickness per model, the architecture itself is not constrained to a specific acquisition protocol. Since our method relies on intra-volume axial references and initially upsampled slices, it can generalize across different slice spacings. In future work, we aim to develop a single unified model capable of handling multiple slice thicknesses and upsampling factors simultaneously.
Importantly, the 5mm slice thickness—used in our primary experiments—is the most common format in real-world clinical CT archives, supporting the immediate practical relevance of our setting.
While our current framework is designed to enhance the overall through-plane resolution in an anatomy-agnostic manner,
task-specific reference selection strategies — such as prioritizing axial slices that share spatial or anatomical relevance with a target organ — may further improve reconstruction quality in organ-focused applications.
Investigating such task-aware or content-guided reference selection mechanisms remains a promising direction for future work.

\section{Conclusion}

In this paper, we introduce a novel anisotropic cross-view texture transfer framework, named ACVTT, to restore through-plane high-resolution details of sparsely sampled CT by utilizing characteristics of anisotropic 3D volume. Specifically, we propose to take in-plane texture details as a reference and transfer them to the through-plane images. A key novelty of our approach lies in using axial slices within the same volume as references, eliminating the need for external datasets. Moreover, we develop a multi-reference non-local attention module (MRNLA) to dynamically extract and transfer texture details from multiple reference slices based on contextual relationships, enhancing the through-plane resolution. Given that multi-reference-based approaches remain under-explored in the domain of reference-based super-resolution, we believe this contribution represents a significant advance in the field.
Our method outperformed state-of-the-art CT slice interpolation models by achieving a 0.5 to 1.0 dB improvement on three large-scale benchmark datasets, including the only real-paired thick-thin CT dataset, thereby establishing a strong new baseline in the field. Additionally, we demonstrated its effectiveness in downstream tasks, such as tumor and kidney segmentation, where ACVTT achieved the highest Dice and NSD scores among competing methods, highlighting its clinical relevance. Comprehensive ablation studies and superior qualitative results further validated the effectiveness of our approach. These contributions underscore ACVTT's potential to advance both the technical and clinical applications of CT slice interpolation.

\bibliography{tmi.bib}
\bibliographystyle{IEEEtran}

\end{document}